\documentclass{article}
\usepackage{spconf, amsmath,amsfonts,epsfig, cite,booktabs, multirow, subcaption, pifont}

\let\OLDthebibliography\thebibliography
\renewcommand\thebibliography[1]{
  \OLDthebibliography{#1}
  \setlength{\parskip}{0pt}
  \setlength{\itemsep}{0pt plus 0.3ex}
}

\begin{document}\sloppy

% Example definitions.
% --------------------
\def\x{{\mathbf x}}
\def\L{{\cal L}}
\def\onedot{.}

\def\eg{\emph{e.g}\onedot} \def\Eg{\emph{E.g}\onedot}
\def\ie{\emph{i.e}\onedot~} \def\Ie{\emph{I.e}\onedot}
\def\cf{\emph{c.f}\onedot} \def\Cf{\emph{C.f}\onedot}
\def\etc{\emph{etc}\onedot} \def\vs{\emph{vs}\onedot}
\def\wrt{w.r.t\onedot} \def\dof{d.o.f\onedot}
\def\etal{\emph{et al}\onedot}

\newcommand{\argmin}{\operatornamewithlimits{arg\,min}}
\newcommand{\cmark}{\ding{51}}%
\newcommand{\xmark}{\ding{55}}%

% Title.
% ------
\title{Turning to A Teacher \\ for Timestamp Supervised Temporal Action Segmentation}
%
% Single address.
% ---------------
\name{Yang Zhao, Yan Song$^{\dagger}$\thanks{$^{\dagger}$Corresponding author.}}
\address{
        Nanjing University of Science and Technology, China \\
         \small \{zhaoyang, songyan\}@njust.edu.cn }

\maketitle

\begin{abstract}
Temporal action segmentation in videos has drawn much attention recently.
Timestamp supervision is a cost-effective way for this task.
To obtain more information to optimize the model, the existing method generated pseudo frame-wise labels iteratively based on the output of a segmentation model and the timestamp annotations.
However, this practice may introduce noise and oscillation during the training, and lead to performance degeneration.
To address this problem, we propose a new framework for timestamp supervised temporal action segmentation by introducing a teacher model parallel to the segmentation model to help stabilize the process of model optimization.
The teacher model can be seen as an ensemble of the segmentation model, which helps to suppress the noise and to improve the stability of pseudo labels. 
We further introduce a segmentally smoothing loss, which is more focused and cohesive,  to enforce the smooth transition of the predicted probabilities within action instances. 
The experiments on three datasets show that our method outperforms the state-of-the-art method and performs comparably against the fully-supervised methods at a much lower annotation cost.
\end{abstract}

\section{Introduction}
\label{sec:intro}
Temporal action segmentation is of vital importance for many real-life applications such as surveillance and interactive robotics.
The goal is to predict frame-wise action labels for a given input video.
This task has got much attention recently, motivating the rapid and remarkable development in the fully-supervised setting (providing frame-wise labels)~\cite{pirsiavash2014parsing,lea2016segmental, lea2017tcn, farha2019mstcn, li2020ms, wang2020bcn, ishikawa2021asrf}.
Due to the high cost of obtaining frame-wise labels, 
some researchers have attempted to solve the problem under weakly-supervised settings, \ie, transcript~\cite{ding2018weakly, richard2018neuralnetwork, li2019weakly} or set~\cite{richard2018action, li2020set, li2021anchor}.
Although the weakly-supervised settings successfully reduce the annotation effort, the performance falls largely behind the fully-supervised counterparts. 
To narrow the gap between them, timestamp supervision has been proposed recently. 
Under this setting, only a single timestamp with its action class (including background) is annotated for each action instance.
Intuitively, timestamp supervision requires a negligible extra cost compared to weakly-supervised settings but provides much more information.

\begin{figure}[t]
  \centering
  \begin{subfigure}[b]{0.65\linewidth}
    \includegraphics[width=\linewidth]{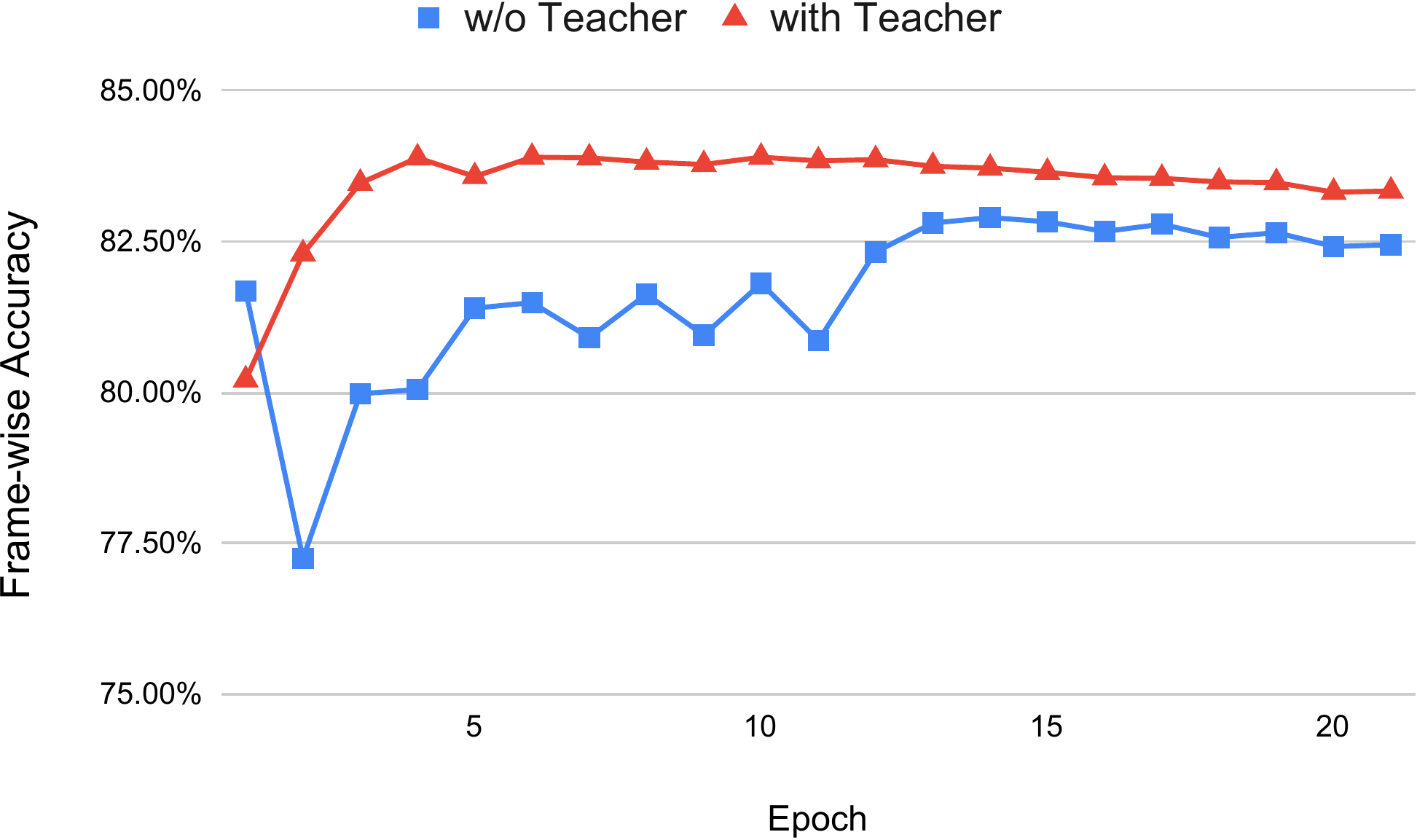}
    \caption{}
  \label{fig:challenge}
  \end{subfigure}
  \begin{subfigure}[b]{0.65\linewidth}
    \includegraphics[width=\linewidth]{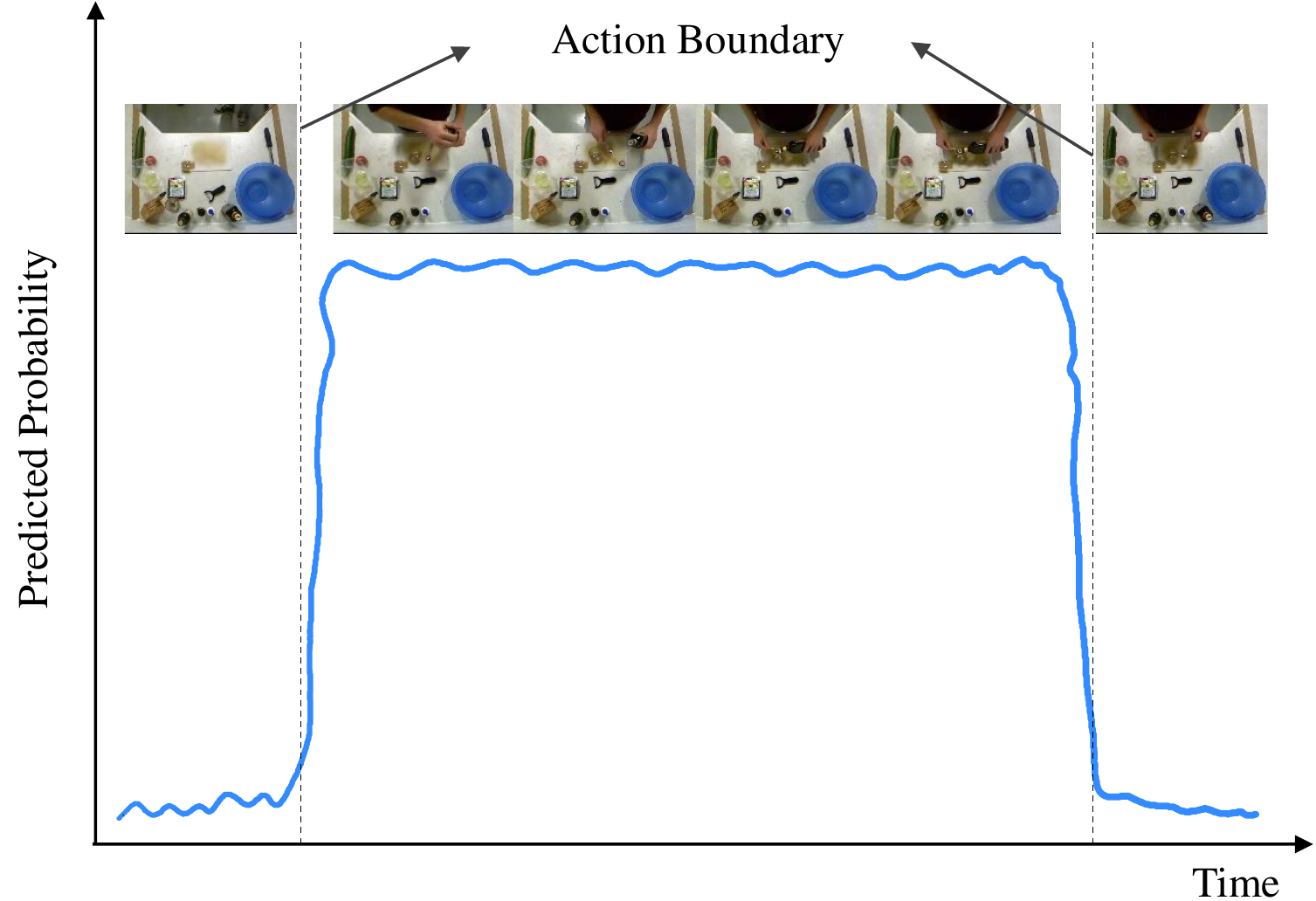}
    \caption{}
  \label{fig:teaser}
  \end{subfigure}
  \caption{
  (a) shows evolution processes of frame-wise accuracy with and without the teacher model on Breakfast~\cite{breakfast}.
  The blue broken line shows that, when the pseudo ground truth is generated iteratively, the predicted frame-wise accuracy is oscillating.
  Contrastively, the red broken line shows that the teacher model is able to suppress the oscillation.
  (b) shows an ideal predicted probability curve for an action instance.
  The predicted probability should have a sudden drop around the action boundary.
  }
\end{figure}
Existing timestamp supervised method~\cite{li2021cvpr} applied a conventional two-step training scheme.
They first adopted a segmentation model designed for full supervision to output frame-level action scores for an input video.
Under the timestamp supervision setting, the model was trained on the timestamp annotations, \ie, sparse set of annotated frames.
Consequently, the model tends to focus on the annotated frames and  those similar to them and ignore the rest frames.
To alleviate this problem, in the second step, they adopted a method to generate pseudo frame-wise labels iteratively which were used as ground truth afterwards.
However, the pseudo ground truth was variational, which may introduce noise and lead to oscillation during the training process~(see the blue broken line in Figure~\ref{fig:challenge}).
Ideally, the predicted probabilities from the segmentation model for an action are supposed to be steadily high within the action but have a sudden drop around the boundaries (see Figure ~\ref{fig:teaser}).
However, in practice, it is observed that they may alternate between high and low values within an action instance.
Although the TMSE loss~\cite{farha2019mstcn} was adopted to encourage smooth transition of the predictions for frames, it was applied on the whole video and all action classes.
Action boundaries were not considered and the constraints on the irrelevant action categories might bring in ambiguity to the model.

In this paper, to address the above questions, we design a new framework for timestamp supervised temporal action segmentation.  
We introduce a teacher model, which is updated by the exponential moving average (EMA) of the segmentation model, in the second training step.  
The EMA can be seen as an ensemble of the current and the earlier versions of the segmentation model, improving the quality and the stability of the teacher model predictions.
In turn, the teacher model  predictions are used to supervise the segmentation model. As a result, the circular dependency helps to improve the stability of pseudo frame-wise labels~(see the red broken line in Figure~\ref{fig:challenge}). 
Additionally, we propose a segmentally smoothing loss that encourages smooth transition within each action instance for predicted probabilities of the corresponding label and  penalizes the low confidence regions surrounded by the high confidence regions. 
Meanwhile, the accumulation in segments makes the constraint more focused and cohesive.
Overall, our contributions are threefold as the following:
\begin{itemize}
  \item  We design a new framework for timestamp supervised temporal action segmentation, 
        where the introduced teacher model helps to maintain the stability of the predictions and pseudo ground truth during training. 
  \item We propose a new segmentally smoothing loss, 
        which encourages smooth transition within an action instance for the predicted probabilities of the labeled action category.
  \item Our approach outperforms state-of-the-art methods on three widely used datasets. 
        Moreover, it even performs comparably against fully-supervised methods.
  
\end{itemize}

\section{Related Work}
% Here, we briefly discuss the work related to our method and cite literature for further reading.

{\bf Fully-supervised temporal action segmentation.}
In order to complete the task of temporal action segmentation, various methods relying on precise frame-wise annotations, \ie full supervision, have been proposed, 
such as grammars\cite{pirsiavash2014parsing}, semi-Markovian model\cite{lea2016segmental} and Temporal Convolution Networks (TCNs)\cite{lea2017tcn}.
Inspired by the success of TCNs, Yazan Abu \etal \cite{farha2019mstcn} introduced MS-TCN, a multi-stage architecture.
It stacks several TCNs, which have dilated convolutions with residual connections. 
There are also some variants \cite{li2020ms,wang2020bcn,ishikawa2021asrf} recently.
However, these methods were trained on frame-wise annotations, which are resource intensive and hard to obtain.

{\bf Weakly-supervised temporal action segmentation.}
To relax the requirement of frame-wise annotations, many works have attempted to explore weaker levels of supervision, such as transcript, set and timestamp. 
In {\bf transcript supervision}, we have access to ordered lists of the action classes occurring in training videos, but their exact action boundaries are unknown.
Ding \etal~\cite{ding2018weakly} proposed Iterative Soft Boundary Assignment (ISBA) to align action sequences and update the network iteratively.
In~\cite{richard2018neuralnetwork,li2019weakly}, they first generated pseudo ground-truth labels for all video frames, and then trained a classifier for frame labeling.
As for {\bf set supervision}, we only have the set of action classes occurring in the training video, without the order and boundaries.
There are many kinds of methods, such as multi-instance learning \cite{richard2018action}, and Viterbi algorithm \cite{li2020set, li2021anchor}.
While both the above levels of supervision reduce the requirement of the annotations, the performance is not superior when compared to the full supervision methods.
To fill this gap,  Li \etal \cite{li2021cvpr} introduced {\bf timestamp supervision} for temporal action segmentation.
They adopted the generated pseudo ground-truth for the subsequent action segmentation by estimating action changes.
Yet the pseudo ground-truth is variational and the frame-wise accuracy tends to oscillate during the training (see Figure~\ref{fig:challenge}). 

% Apart from fully- and weakly-supervised temporal action segmentation, a few works have explored {\bf unsupervised learning}\cite{kukleva2019unsupervised,li2021action}. 
% However, it is outside the scope of this paper, and our focus is on timestamp supervised temporal action segmentation.

{\bf Mean Teacher},
a method that is traditionally adopted in the semi-supervised learning tasks, follows a student-teacher network~\cite{meanteacher}.
It obtains a better teacher model by moving the average of the student model weights without additional training. 
In effect, the teacher model is an ensemble of the student model, enabling itself to learn more abstract invariance and yield more accurate predictions.

\section{Method}
\begin{figure*}[t]
  \centering
  \includegraphics[clip=true, width=1.0\textwidth]{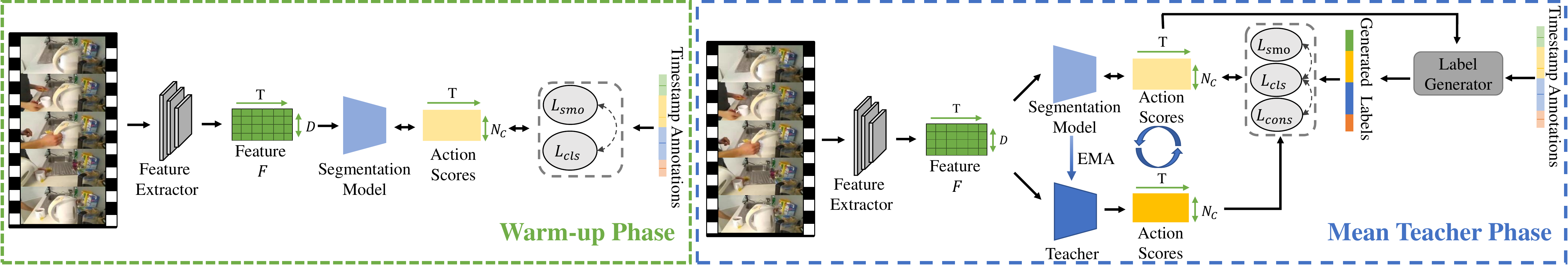}
  \caption{
  The pipeline of our proposed framework consists of two steps: a warm-up phase and a mean teacher phase.
  In the warm-up phase, we train a segmentation model on the timestamp annotations.
  In the mean teacher phase, we introduce a teacher model  parallel to the segmentation model to the framework.
  It is obtained by the EMA of the segmentation model, 
  while the segmentation model is updated by the supervision from the generated pseudo labels, which are generated by a label generator, and the teacher model predictions.
  This forms a circular dependency.  %that help to maintain stability and obtain better performance.
}
 
  \label{fig:intro}
\end{figure*}
\subsection{Problem Definition}
Given a video $X=\{x_1, \dots, x_T\}$ with $T$ frames, 
the goal of the temporal action segmentation task is to predict a sequence of the frame-wise labels $[\tilde{a}_1, \dots, \tilde{a}_T]$.
Unlike full supervision, timestamp supervision provides a set of annotated timestamps $A=\{a_{t_1}, \dots, a_{t_n} ,\dots, a_{t_N}\}$ with each video, where $N$ is the number of the annotated timestamps. 
Especially, for each action instance within the video $X$, we have access to one frame labeled with timestamp $t_n \in [1, T]$ and action class $a_{t_n}$, where $n$ denotes the $n$-th action instance and $n \in [1, N]$.
Although a video may contain multiple action instances, the number of labeled timestamps $N$ is much smaller than the number of frames $T$, \ie, $N \ll T$.
This reveals the sparse nature of  timestamp supervision.

\subsection{Overview}
We propose a framework for timestamp supervised temporal action segmentation including a warm-up phase and a mean teacher phase.
As shown in Figure~\ref{fig:intro}, we use the feature  $F \in \mathbb{R}^{T \times D}$ extracted by I3D~\cite{2017i3d} as input to our framework.
We first train a segmentation model on the timestamp annotations, which is denominated as the warm-up phase.
Then we adopt a label generator to generate pseudo frame-wise labels to alleviate the problem of the sparse nature of timestamp supervision.
However, this will introduce oscillation that results in performance degeneration.
To alleviate this problem, inspired by Mean Teacher~\cite{meanteacher},  we propose a teacher model parallel to the segmentation model, which is updated by the EMA of the segmentation model.
In turn, the teacher model predictions are used to update the segmentation model.
This forms a circular dependency that helps the framework maintain stability and obtain better performance, which is denominated as the mean teacher phase.
The combination of the segmentation and the teacher models is similar to a real-life scenario.
A student, who encounters problems, will turn to a teacher.
The process of solving problems helps the student enrich knowledge and the teacher gather experience.

\subsection{Warm-up Phase}
\begin{figure}[tb]
  \centering
  \includegraphics[clip=true, width=0.3\textwidth]{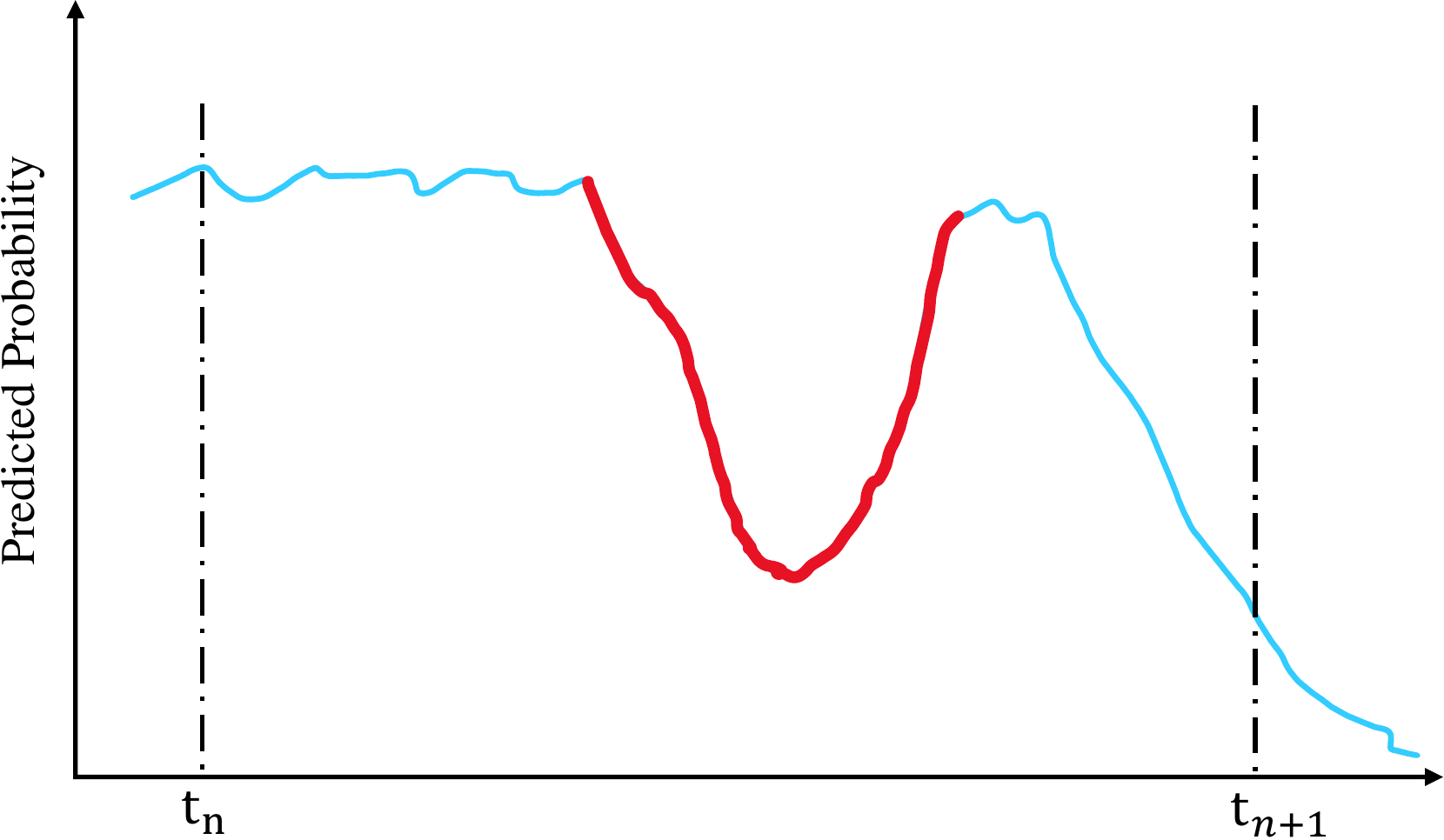}
  \caption{The red part of the curve, namely the low confidence region, should have the similar predicted probabilities as the adjacent blue curve.
          The proposed segmentally smoothing loss will penalize the frames in the red part.
          }
  \label{fig:smooth}
\end{figure}
In the warm-up phase, we only have access to a sparse set of annotated frames.
To train a segmentation model, we apply a cross entropy loss as the  partial classification loss 
\begin{equation}\label{cls-loss}
    L_{cls} = \frac{1}{N} \sum_{n=1}^N{-log(y_{t_n,c})},
\end{equation}
where $y_{t_n,c}$ denotes the predicted probability for the target action class $c$ at annotated timestamp $t_n$.
It ignores all frames that are not annotated.
Note that the predicted probability is the Softmax output of the action scores $S \in \mathbb{R}^{T \times N_c}$, where $N_c$ indicates the number of action classes.

Due to the sparse nature of timestamp supervision, it is observed that the segmentation model only learns from each action instance the annotated frame and other informative frames which have similar features to the annotated frame.
Intuitively, the frames between the annotated frames and the informative frames should have similar predicted probability, and these frames should be predicted to have high confidence.
Based on this, we propose a segmentally smoothing loss
\begin{equation}\label{smoothing}
\begin{split}
    & L_{smo} =  \sum_{\substack{a_{t_n} \in A}}{\left(\frac{1}{T'}
                                            \sum_{t=t_{n-1}+1}^{t_{n+1}}{
                                                \delta_{a_{t_n},t }
                                                }\right)
                                            } \\
    s.t. & \\ 
    & \delta_{a_{t_n},t}=\min(\tau,|\log{y_{t,a_{t_n}}-\log{y_{t-1,a_{t_n}}}}|),
\end{split}
\end{equation}
where $y_{t,a_{t_n}}$ refers to the predicted probability for the annotated action class $a_{t_n}$ at time $t$, 
$T'$ denotes the number of frames between the annotated frames $t_{n-1}$ and $t_{n+1}$, and $\tau$ is a hyperparameter.
The segmentally smoothing loss encourages smooth transition of the frames between high confidence frames and therefore suppresses the low confidence regions surrounded by high confidence frames (see Figure~\ref{fig:smooth}).
It helps the segmentation model to learn from not only the annotated and the informative frames but also those frames in low confidence regions.
Unlike TMSE~\cite{farha2019mstcn}, which encourages smooth transition of the predicted probabilities for all frames in an input video and all action classes, 
the proposed loss accumulates loss segmentally to encourage smooth transition for the labeled action categories within action instances in the input video.
So it is more focused and cohesive.

The overall loss function for the segmentation model in the warm-up phase is a weighted sum of the above two losses
\begin{equation}\label{aux-loss}
    \mathcal{L} = L_{cls} + \alpha L_{smo},
\end{equation}
where $\alpha$ is a hyperparameter to balance the contribution of each loss.

\subsection{Mean Teacher Phase}
Although the smoothing loss is able to help pay attention to the low confidence parts of an action instance, it still ignores many frames without annotations.
Following~\cite{li2021cvpr}, we apply a label generator to generate pseudo frame-wise labels by detecting action changes.
However, we find that the generated pseudo frame-wise labels are not stable and cause oscillation during the training process.
To alleviate this problem, we introduce a teacher model parallel to the segmentation model to help stabilize the system, 
which has the same architecture as the segmentation model.
We update the teacher model by the EMA of the segmentation model.
The EMA is formed by an ensemble of the current version and those earlier versions of the segmentation model.
At $i$-th iteration, it is defined as
\begin{equation}\label{ema}
  \overline{\theta}_{i}=\left\{
          \begin{array}{ll}
              (1-\frac{1}{i}) \overline{\theta} _{i-1} + \frac{1}{i} \theta_i & 1-\frac{1}{i} < \lambda,  \\
              \lambda \overline{\theta} _{i-1} + (1-\lambda)\theta_i & otherwise, \\
          \end{array}
  \right.
\end{equation}
where $\theta$ and $\overline{\theta}$ are the weights of the segmentation model and the teacher model  respectively, and $\lambda$ is a hyperparameter.
At the initial training iterations, the teacher model is a bare model without training.
To help the teacher model learn quickly, we use absolute average~\cite{chen2021seminar} instead of EMA when $1-\frac{1}{i} < \lambda$.

To make the consistency constraint between the segmentation and the teacher models, 
we adopt a frame consistency loss $L_{cons}$ between their predictions, in the form of mean square error (MSE)\cite{meanteacher}.
\begin{equation}\label{cons-loss}
    L_{cons} = -\frac{1}{T} \sum_{t\in T}\|f(F, \theta) -f(F, \overline{\theta})\|^2 ,
\end{equation}
where $f(\cdot)$ denotes the predicted probabilities of a model.
With this loss, the segmentation model learns from the teacher model, thereby forming a circular dependence as shown in Figure~\ref{fig:intro}. 
It helps the two models and the generated pseudo labels stay steady.

The overall loss function for the segmentation model in the mean teacher phase is a weighted sum of three losses
\begin{equation}\label{mt-loss}
    \mathcal{L} = L_{cls} + \alpha L_{smo} + \beta L_{cons},
\end{equation}
where $\beta$, akin to $\alpha$, is a hyperparameter. % to balance the contribution of each loss.
For simplicity, we use $L_{cls}$ to denote both pseudo classification and partial classification losses. 
% While they are in the same form of cross-entropy loss, the pseudo classification loss  is supervised by the pseudo frame-wise labels defined as 
While they are in the same form of cross-entropy loss, the pseudo classification loss
\begin{equation}\label{p-cls-loss}
    L_{cls} = \frac{1}{T} \sum_{t=1}^T{-log(y_{t,c})}
\end{equation}
is supervised by the pseudo frame-wise labels.
To generate the pseudo frame-wise labels, 
% we regard the timestamp that minimizes the energy function in~\cite{li2021cvpr} as the action change between any two annotated frames and assign to each frame located between an annotated frame and an estimated action change the same label. % as the annotated frame.
we regard the timestamp that minimizes the energy function in~\cite{li2021cvpr} as the action change between any two annotated frames.
And then, we assign the same label to each frame located between an annotated frame and an estimated action change.
Different from~\cite{li2021cvpr}, we utilize action scores as the input feature to the energy function, rather than the output of the penultimate layer in the segmentation model.
It is worth mentioning that we find that there is no benefit to perform forward-backward estimating as in~\cite{li2021cvpr} experimentally, 
so we only adopt the stamp-to-stamp version of it.
Note that the above loss function is only computed for the segmentation model, and no gradient is computed on the teacher model.

\section{Experiments}

\subsection{Datasets and Evaluation Metrics}
We evaluate our method on three public datasets, 
namely Georgia Tech Egocentric Activities (GTEA)\cite{gtea}, 50Salads\cite{50salads} and Breakfast\cite{breakfast}.

{\bf GTEA} contains 28 videos, 11 action classes and 31,222($\approx$ 31K) frames.
It covers 7 daily activities which were performed by 4 actors in a kitchen.
For evaluation, we use fourfold cross-validation and report the average as in \cite{farha2019mstcn}.

{\bf 50Salads} contains 50 videos with 17 action classes and 518,411($\approx$ 520K) frames.
On average, each video contains 20 action instances and is 6.4 minutes long.
These action instances were performed by 25 actors and recorded from the top view, where each actor prepared a mixed salad twice.
We use fivefold cross-validation for evaluation and report the average following \cite{farha2019mstcn}.

{\bf Breakfast} is the largest among the three datasets.
It contains 1712 third-person view videos related to breakfast preparation activities.
These videos were recorded in 18 different kitchens.
Overall, there are 48 different action classes and roughly 3.6M frames.
Each video contains 6 action instances on average.
To evaluate, we use the standard 4 splits proposed in \cite{breakfast} and report the average.

Following \cite{farha2019mstcn}, for all the above datasets, we use the temporal resolution of 15 fps.
We use the timestamp annotations that Li \etal \cite{li2021cvpr} provide.

{\bf Evaluation Metrics.}
Following the standard protocol of temporal action segmentation\cite{li2021cvpr,lea2016segmental},
we report frame-wise accuracy (Acc), 
segmental edit distance (Edit) and segmental F1 scores at intersection over union (IoU) thresholds $10\%$, $25\%$ and $50\%$.

\subsection{Implementation Details}
We take the I3D~\cite{2017i3d} features extracted by~\cite{farha2019mstcn} as the input to our framework and the MS-TCN~\cite{farha2019mstcn} as our segmentation model.
Following \cite{li2021cvpr}, we use two parallel stages for the first stage with kernel size 3 and 5, 
change the number of layers in the first stage from 10 to 12,  and pass the sum of both outputs to the next stages.
We train our model for 50 epochs with Adam optimizer.
Out of the 50 epochs, the warm-up phase takes 30 epochs, while the mean teacher phase takes the remaining 20 epochs.
We use the learning rate 0.0001 for Breakfast and  0.0005  for GTEA and 50Salads, then multiply them by 0.1 every 40 epochs. 
For the hyperparameters, they are determined empirically.
We set $\lambda$ to 0.9, 0.99 and 0.999 for GTEA, 50Salads and Breakfast respectively.
Both $\tau$ and $\alpha$ are set to 1.
To help the teacher model learn progressively, we increase $\beta$ linearly from 0 to 0.5 during the 20 epochs.

During the inference time, we take the output of the teacher model as the final predictions~\cite{meanteacher}, and classify each frame as the label with maximum probability.
\subsection{Comparison with the State-of-the-art}

\begin{table}[tb]
\begin{center}
\caption{Comparison with the state-of-the-art on GTEA.}
\label{tab:gtea}
\resizebox{0.8\columnwidth}{!}{
\begin{tabular}{c|l|ccc|c|c}
\toprule
\multirow{2}{*}{Supervision} &
\multicolumn{1}{c|}{\multirow{2}{*}{Method}} &
\multicolumn{3}{c|}{F1@IoU (\%)} & \multirow{2}{*}{Edit} & \multirow{2}{*}{Acc}\\
        &  & 10  & 25  & 50  &  \\
      \midrule\midrule

\multirow{4}{*}{Full}
        & MS-TCN\cite{farha2019mstcn}  & 85.8  & 83.4  & 69.8  & 79.0 & 76.3   \\
        & MS-TCN++\cite{li2020ms}  & 88.8  & 85.7  & 76.0  & 83.5 & 80.1   \\
        & BCN\cite{wang2020bcn}  & 88.5  & 87.1  & 77.3  & 84.4 & 79.8   \\
        & ASRF\cite{ishikawa2021asrf}  & 89.4  & 87.8  & 79.8  & 83.7 & 77.3   \\
        \midrule

\multirow{2}{*}{Timestamp} 
       & Li~\etal\cite{li2021cvpr}  & 78.9  & 73.0  & 55.4  & 72.3 & 66.4 \\
       & Ours  & \textbf{84.3}  & \textbf{81.7}  & \textbf{64.8}  & \textbf{79.8} & \textbf{74.4} \\
    \bottomrule
\end{tabular}
}
\end{center}
\end{table}
\begin{table}[tb]
\begin{center}
\caption{Comparison with the state-of-the-art on 50Salads.}
\label{tab:50}
\resizebox{0.8\columnwidth}{!}{
\begin{tabular}{c|l|ccc|c|c}
\toprule
\multirow{2}{*}{Supervision} &
\multicolumn{1}{c|}{\multirow{2}{*}{Method}} &
\multicolumn{3}{c|}{F1@IoU (\%)} & \multirow{2}{*}{Edit} & \multirow{2}{*}{Acc}\\
        &  & 10  & 25  & 50  &  \\
      \midrule\midrule

\multirow{4}{*}{Full}
        & MS-TCN\cite{farha2019mstcn}  & 76.3  & 74.0  & 64.5  & 67.9 & 80.7   \\
        & MS-TCN++\cite{li2020ms}  & 80.7  & 78.5  & 70.1  & 74.3 & 83.7   \\
        & BCN\cite{wang2020bcn}  & 82.3  & 81.3  & 74.0  & 74.3 & 84.4   \\
        & ASRF\cite{ishikawa2021asrf}  & 84.9  & 83.5  & 77.3  & 79.3 & 84.5   \\
        \midrule

\multirow{2}{*}{Transcript} 
       & NN-Viterbi\cite{richard2018neuralnetwork}  & -  & -  & -  & - & 49.4  \\ 
       & CDFL\cite{li2019weakly}  & -  & -  & - & -  & 54.7  \\
       \midrule

% \multirow{2}{*}{Unsupervised}
        % & CTE\cite{kukleva2019unsupervised}  & -  & -  & 26.4  & - & 41.8   \\
        % & ASAL\cite{li2021action}  & -  & -  & 37.9  & - & 52.5   \\
        % \midrule

\multirow{2}{*}{Timestamp} 
       & Li~\etal\cite{li2021cvpr}  & 73.9  & 70.9  & 60.1  & 66.8 & 75.6  \\
       & Ours  & \textbf{78.5}  & \textbf{75.5}  & \textbf{63.4}  & \textbf{71.8} & \textbf{77.7} \\
    \bottomrule
\end{tabular}
}
\end{center}
\end{table}
\begin{table}[tb]
\begin{center}
\caption{Comparison with the state-of-the-art on Breakfast.}
\label{tab:bf}
\resizebox{0.8\columnwidth}{!}{
\begin{tabular}{c|l|ccc|c|c}
\toprule
\multirow{2}{*}{Supervision} &
\multicolumn{1}{c|}{\multirow{2}{*}{Method}} &
\multicolumn{3}{c|}{F1@IoU (\%)} & \multirow{2}{*}{Edit} & \multirow{2}{*}{Acc}\\
        &  & 10  & 25  & 50  &  \\
      \midrule\midrule

\multirow{4}{*}{Full}
        & MS-TCN\cite{farha2019mstcn}  & 52.6  & 48.1  & 37.9  & 61.7 & 66.3   \\
        & MS-TCN++\cite{li2020ms}  & 64.1  & 58.6  & 45.9  & 65.6 & 67.6   \\
        & BCN\cite{wang2020bcn}  & 68.7  & 65.5  & 55.0  & 66.2 & 70.4   \\
        & ASRF\cite{ishikawa2021asrf}  & 74.3  & 68.9  & 56.1  & 72.4 & 67.6   \\
        \midrule

\multirow{2}{*}{Transcript} 
       & NN-Viterbi\cite{richard2018neuralnetwork}  & -  & -  & -  & - & 43.0  \\ 
       & CDFL\cite{li2019weakly}  & -  & -  & - & -  & 50.2  \\
       \midrule

\multirow{3}{*}{Set}
        & ActionSet\cite{richard2018action}  & -  & -  & -  & - & 23.3   \\
        & SCV\cite{li2020set}  & -  & -  & -  & - & 30.2   \\
        & ACV\cite{li2021anchor}  & -  & -  & -  & - & 33.4   \\
        \midrule

% \multirow{2}{*}{Unsupervised}
%         & CTE\cite{kukleva2019unsupervised}  & -  & -  & 26.4  & - & 41.8   \\
%         & ASAL\cite{li2021action}  & -  & -  & 37.9  & - & 52.5   \\
%         \midrule

\multirow{2}{*}{Timestamp} 
       & Li~\etal\cite{li2021cvpr}  & 70.5  & 63.6  & 47.4  & 69.9 & 64.1  \\
       & Ours  & \textbf{73.1}  & \textbf{66.5}  & \textbf{49.4}  & \textbf{72.6} & \textbf{64.6} \\
    \bottomrule
\end{tabular}
}
\end{center}
\end{table}

We compare our method with recent approaches under different levels of supervision. 
The results are presented in Tables \ref{tab:gtea}, \ref{tab:50} and \ref{tab:bf}.
As shown in these tables, our method outperforms  the state-of-the-art timestamp supervised approach with a large margin
on all datasets (up to 9.4\% for the F1 score at 50\% IoU threshold on GTEA).
At the same time, our method exceeds the approaches under weaker levels of supervision in the form of transcript or set at a comparable annotation cost.
Furthermore, our method even performs comparably against the fully-supervised methods in terms of F1 scores and Edit at a much lower annotation cost.
Despite the lack of the information of the action boundary, 
% our method achieves $96\% \sim 97\%$ of the frame-wise accuracy of the full supervision method MS-TCN~\cite{farha2019mstcn}.
our method achieves about $97\%$ frame-wise accuracy of the full supervision method MS-TCN~\cite{farha2019mstcn}.

\subsection{Ablation Study}
\label{sec:exp}

\begin{table}[t]
\centering
\caption{
Ablation study on GTEA and 50Salads. $\mathbf{T}$ denotes the teacher model.
}
\resizebox{0.8\columnwidth}{!}{
\begin{tabular}{l|ccc|ccc|c|c}
    \toprule
    \multirow{2}{*}{Dataset} & \multirow{2}{*}{$\mathbf{T}$} & \multirow{2}{*}{$L_{smo}$} & \multirow{2}{*}{$L_{T-MSE}$} 
    & \multicolumn{3}{c|}{F1@IoU (\%)}  & \multirow{2}{*}{Edit}  & \multirow{2}{*}{Acc} \\
       & & &  & 10  & 25  & 50  &  &   \\
    \midrule\midrule
       \multirow{4}{*}{GTEA} 
                &  &\cmark & & 82.0 & 77.0 & 58.1  & 76.9  & 71.7 \\
                & \cmark  & & &  78.3  & 75.2  & 55.4 & 70.8 & 67.0 \\
                & \cmark  & & \cmark &  79.0 & 72.6 & 49.1 & 75.3 & 59.6 \\
                & \cmark  & \cmark & &  \textbf{84.3}  & \textbf{81.7}  & \textbf{64.8} & \textbf{79.8}  & \textbf{74.4} \\
    \midrule
       \multirow{4}{*}{50Salads} 
                & & \cmark &  & 77.8 & 75.1 & 63.1  & 69.6  & 77.7 \\
                & \cmark & & &  62.5  & 58.6  & 45.8    & 54.0  & 72.1 \\
                & \cmark & & \cmark &  75.7  & 71.8  & 60.0 & 67.5 & 76.9 \\
                & \cmark & \cmark & &  \textbf{78.5}  & \textbf{75.5}  & \textbf{63.4} & \textbf{71.8}  & \textbf{77.7} \\
    \bottomrule
\end{tabular}
}
\label{tab:abla_mod}
\end{table}
{\bf Study on Teacher Model.}
To validate the efficacy of the proposed teacher model, we compare the results of  our framework with and without the teacher model. We adopt the conventional two-step training scheme for the latter one.
As shown in Table~\ref{tab:abla_mod}, the framework with the teacher model outperforms the one without it.
We also present the evolution processes of frame-wise accuracy during the mean teacher phase for the two settings in Figure~\ref{fig:challenge}, 
and it demonstrates that  the teacher model is able to maintain stability during training and achieves favorable performance.
On the contrary, the conventional two-step training scheme (without the teacher model) introduces a severe oscillation which results in performance degeneration.

{\noindent {\bf Study on Smoothing Loss.}
\begin{figure}[tb]
  \centering
  \includegraphics[clip=true, width=0.35\textwidth]{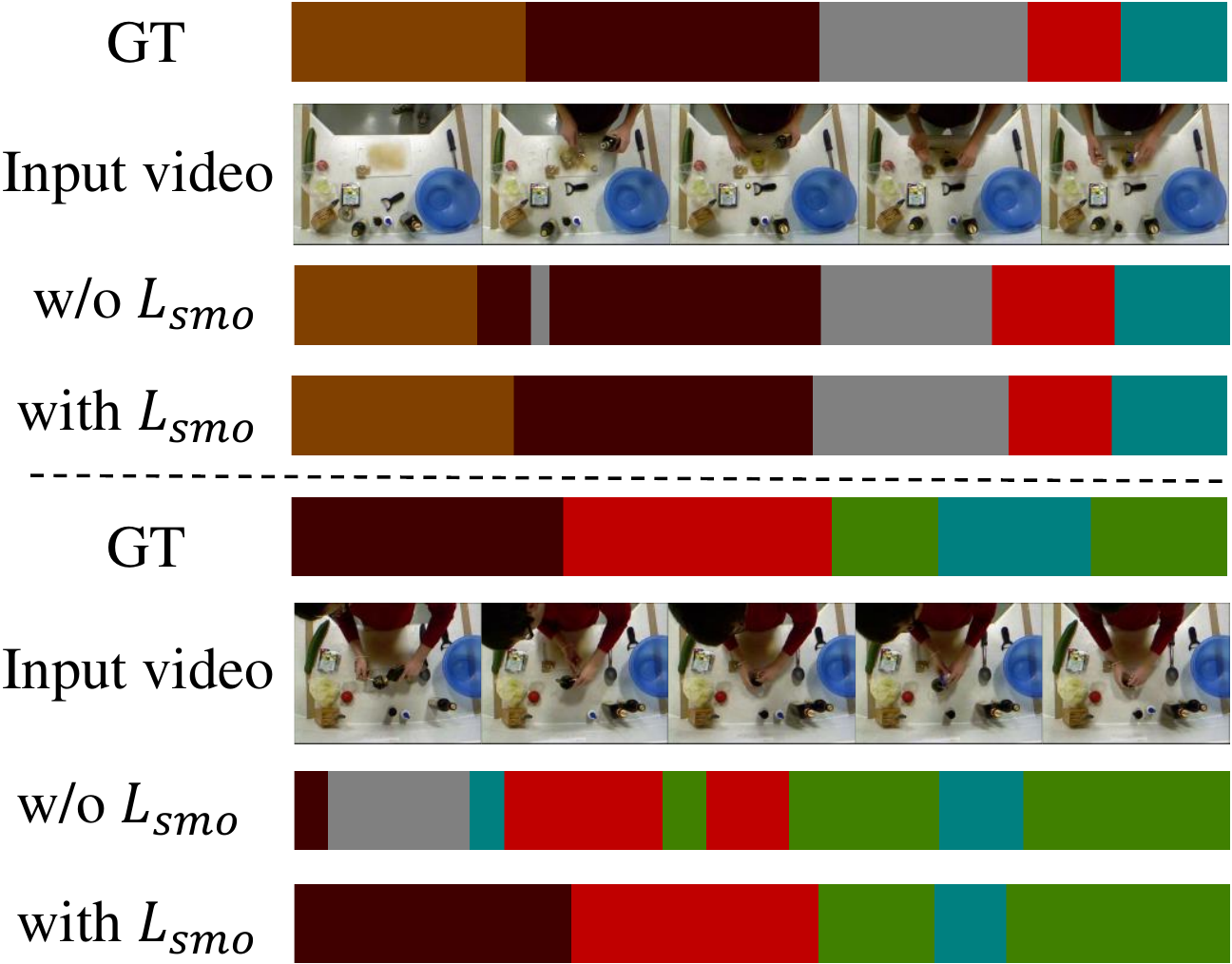}
  \caption{
            Qualitative results of our method with and without the proposed segmentally smoothing loss on 50Salads.
          }
  \label{fig:vis}
\end{figure}
We also verify the effectiveness of the proposed segmentally smoothing loss. We list the results of three runs on GTEA and 50Salads in Table \ref{tab:abla_mod}, 
including the one without any smoothing loss, the one with TMSE~\cite{farha2019mstcn} and the one with the proposed segmentally smoothing loss.
As analyzed earlier, the method without smoothing loss suffers from a severe over-segmentation error, which is indicated by the low F1 scores and Edit.
The proposed segmentally smoothing loss outperforms the TMSE by $5\%\sim 9\%$ and $3\%\sim 4\%$ on F1 scores on the two datasets respectively.
Furthermore, to present the impact of the proposed segmentally smoothing loss visually, we show two representative examples from 50Salads in Figure~\ref{fig:vis}.

\section{Conclusion}
In this paper, we propose a new framework for timestamp supervised temporal action segmentation by introducing a teacher model to help the segmentation model yield more accurate predictions. 
With the help of the teacher model, the problem of the instability of the pseudo labels can be relieved.
We also propose a segmentally smoothing loss that encourages smooth transition within each action instance.
The loss penalizes the low confidence regions that are surrounded by high confidence regions.
Our method outperforms the state-of-the-art method on three datasets and further narrows the performance gap between timestamp supervision and full supervision.

\small{\noindent\textbf{Acknowledgements:} This work is supported by the National Key RD Program of China (No. 2018AAA0102002), the National Natural Science Foundation of China (Grant No. 61672285).}

\bibliographystyle{IEEEbib}
\bibliography{mini}
% \bibliography{icme2020template}

\end{document}